\documentclass[letterpaper, 10 pt, conference]{ieeeconf}  

\IEEEoverridecommandlockouts                              

\usepackage{url}
\usepackage{cite}
\usepackage{xcolor}
\usepackage{amsfonts}
\usepackage{graphicx}
\usepackage{amsmath}
\usepackage{booktabs}
\usepackage{makecell}
\usepackage{float}
\usepackage{tabularx}
\usepackage{caption}
\usepackage[font=small]{caption}

\title{\LARGE \bf
CEER: Compliant End-Effector and Root Control as a Unified Interface for Hierarchical Humanoid Loco-Manipulation}

\author{
Xinyuan Luo, Xingrui Chen, Xunjian Yin, Hongxuan Wu,\\
Boxi Xia, Zhuoqun Chen, Jinzhou Li, Boyuan Chen, Xianyi Cheng%
\thanks{This work is supported by DARPA TIAMAT program under award HR00112490419 and ARO under award W911NF2410405.}%
\thanks{All authors are with the Department of Mechanical Engineering and Materials Science, Duke University, Durham, NC 27708, USA.}%
}

\begin{document}
\maketitle
\thispagestyle{empty}
\pagestyle{empty}
\bstctlcite{IEEEexample:BSTcontrol}

\begin{abstract}


Humanoid robots have achieved impressive locomotion performance, yet contact-rich and long-horizon manipulation remains a major bottleneck. 
Manipulation is inherently contact-rich and demands compliant whole-body control for stable interaction, while its diversity and long-horizon nature favor modular, planner-compatible interfaces over joint-space tracking.

We propose CEER, a compliant end-effector–root (EE-root) control abstraction for modular humanoid loco-manipulation within a hierarchical planning framework. CEER enables compliance-aware whole-body control in an interpretable task space defined by root motion commands and end-effector pose targets, and supports plug-and-play integration with heterogeneous high-level planners. A teacher–student framework is adopted to distill a general motion-tracking controller into a low-level policy that consumes only EE-root commands.

We further construct a hierarchical system that integrates heterogeneous planners and task modules through the EE-root interface, enabling diverse manipulation tasks without retraining the underlying whole-body policy. Experiments in simulation and on hardware demonstrate 3.3\,cm end-effector tracking accuracy with substantially reduced jerk compared to baselines, stable contact-rich manipulation under teleoperation, and up to 70\% success in simulated single-object loco-manipulation tasks within a room-scale environment.
These results indicate that compliant EE-root control provides a practical abstraction for humanoid loco-manipulation, enabling modular and scalable integration of diverse skills. Project page: \url{https://robotproject8.github.io/ceer_page/}



\end{abstract}

\section{INTRODUCTION}


For humanoid robots to operate effectively in everyday human environments, they must safely perform contact-rich manipulation while adapting to a wide range of tasks. Humanoid robots today demonstrate advanced locomotion behaviors. Powered by motion-capture data and imitation learning \cite{deepmimic}, modern humanoids are now capable of performing highly dynamic motions such as cartwheels \cite{beyondmimic} and backflips \cite{omniretarget}. By inputting humanoid robot whole-body target joint angles, those policies are able to output physically feasible joint commands precisely and robust to disturbances. Despite their strong tracking performance, motion imitation policies are typically blind and primarily optimized for locomotion and ground interaction. When extended to object manipulation, they often require task-specific retraining or retargeting to accommodate contact dynamics~\cite{hdmi, omniretarget}. 
\begin{figure}[htbp]
\begin{center}
\includegraphics[scale=0.45]
{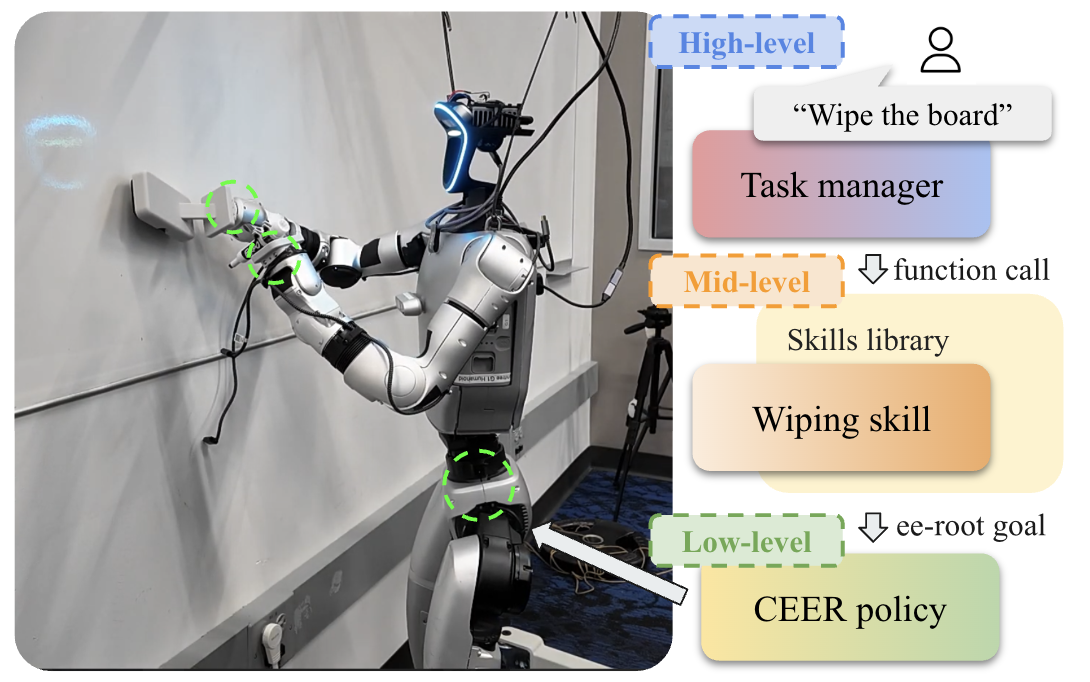}
\end{center}
\vspace{-3mm}
\caption{
CEER defines a compliant end-effector–root (EE-root) control abstraction for whole-body humanoid loco-manipulation.
Built upon this unified interface, we construct a three-layer hierarchical system in which an LLM-based task manager invokes modular mid-level skills that generate EE-root goals for the low-level controller.
This design highlights the modularity and extensibility enabled by the unified command interface.
}
\label{fig: teaser}
\vspace{-5mm}
\end{figure}

More fundamentally, joint-tracking methods are difficult to fully automate the perception–planning–execution pipeline. First, these methods require full-body joint targets as input. As a result, the planner must generate high-dimensional joint trajectories directly from perception and task descriptions. This is extremely challenging given the limited availability of whole-body data. Second, joint-level commands are inherently embodiment-specific, which prevents direct transfer across robots with different kinematic structures and reducing data efficiency. Finally, pure motion tracking lacks explicit compliance modulation, which is crucial for safe and robust interaction with objects and humans.


To address this challenge, we seek an effective task-level representation for whole-body loco-manipulation. In contact-rich manipulation, interaction primarily occurs at the end-effectors, making end-effector (EE) targets a natural task-space abstraction. Extending this principle to loco-manipulation, we incorporate root motion to account for locomotion. This EE-root representation significantly reduces the planner’s output dimensionality while unifying task-level commands, enabling efficient reuse and integration of diverse planning modules. The remaining challenge is to bridge this EE-root abstraction with dynamically feasible whole-body control.

To this end, we propose CEER, a EE-root control abstraction for whole-body humanoid loco-manipulation. 
With CEER, we achieve two essential capabilities for effective and general humanoid loco-manipulation:
(1) compliance for contact-rich tasks, and 
(2) interpretable task-level command inputs compatible with high-level planning.  Importantly, CEER enables plug-and-play integration of heterogeneous planners and skills without retraining the whole-body controller.
 

We first develop a unified compliant controller to bridge EE-root commands with whole-body execution. We train a compliant whole-body motion tracking policy using AMASS \cite{amass}, considering compliance as an impedance behavior following prior works \cite{unifiedforce, gentlehumanoid} and incorporating external disturbances during training \cite{gentlehumanoid}. This tracking policy serves as a teacher to distill a student policy that operates purely on EE-root commands, thereby decoupling morphology-specific joint targets from task-level control inputs.
We further introduce a three-layer hierarchical architecture comprising a high-level skill manager, modular mid-level skills, and the unified CEER low-level controller, illustrating how the EE-root abstraction supports modular and planner-agnostic integration. In this structure, mid-level skills include teleoperation interfaces, scripted locomotion and manipulation primitives, and learnable diffusion-based skills. All mid-level skills produce unified EE-root commands as direct input to the low-level policy CEER, which is essential for the plug-and-play integration and task generalizability. 

We evaluate CEER from both control and system perspectives. First, we quantify EE pose tracking accuracy and motion smoothness. Second, we validate contact-rich manipulation behaviors on real hardware via teleoperation. Finally, we demonstrate long-horizon hierarchical task execution in simulation, highlighting the modularity and scalability of the proposed interface.

In summary, our contributions are:

\begin{itemize}
    \item We propose CEER, a compliant end-effector–root (EE-root) control abstraction that defines a unified execution interface for humanoid loco-manipulation, implemented as a whole-body policy, enabling task-level command specification independent of morphology-specific joint targets.
    
    \item We design a hierarchical three-layer framework that decouples high-level planning from low-level control through a unified planner-compatible interface, allowing plug-and-play integration of heterogeneous planners.
    
    \item We demonstrate the modularity and scalability of CEER through long-horizon loco-manipulation tasks in a room-scale environment, integrating natural language planning, teleoperation, and learnable mid-level skills without retraining the low-level policy.
\end{itemize}













\section{Related Work}

\subsection{Learning Humanoid Whole-Body Control}
Whole-body control for humanoid is difficult due to nonlinear underactuated dynamics and complex multi-contact interactions, which complicates stable loco-manipulation under actuation limits and state estimation noise. Recent advances in reinforcement learning and sim-to-real transfer have enabled complex whole-body skills by training policies in simulation with task-specific rewards and motion imitation signals \cite{deepmimic}. Building on this direction, data-driven imitation at scale improves expressive tracking on hardware through large motion datasets and retargeting pipelines, including teleoperation-derived data for human-to-humanoid control \cite{exbody,exbody2,h2o,omnih2o}. Other works further improve stability and manipulation accuracy by conditioning locomotion on predicted upper-body motion, or by introducing physics-aware processing to better handle highly dynamic movements \cite{mobile-tv,kungfubot}. Recent trends incorporate generative models for skill composition \cite{beyondmimic} and visual conditioning with hierarchical keypoint control for loco-manipulation \cite{visualmimic}. Despite these advances, most approaches still prioritize tracking fidelity and expressiveness over contact-rich interaction behaviors.

\subsection{Compliance and Force-adaptive Loco-manipulation}
Recent works incorporate compliance and force adaptation into whole-body humanoid control by injecting impedance-like behavior into motion tracking \cite{gentlehumanoid}, augmenting data with compliant variants \cite{softmimic}, or learning stiffness modulation for contact-rich tasks such as wiping and door opening \cite{chip}.
Force-adaptive loco-manipulation has also been studied via curriculum learning and body-part decomposition for payload transport and cart pulling \cite{falcon}. Related position--force and impedance-reference tracking frameworks for legged systems improve contact robustness without force sensors and can extend to loco-manipulators \cite{unifiedforce,facet}.
However, these methods often expose tracking-centric interfaces (e.g., joint trajectories, dense full-body references, or task-specific force/impedance setpoints), whereas modular planners and mid-level skills typically produce sparse task-level goals (root navigation and end-effector poses). This mismatch can require additional retargeting, interface adapters, or retraining for system integration and cross-embodiment reuse. In contrast, our EE-root abstraction unifies root motion commands and end-effector pose targets in a compact task space, enabling plug-and-play composition of planners and skills while retaining compliant whole-body behavior under contact.




\subsection{Humanoid Teleoperation and Control Interfaces}
Human-to-humanoid interfaces can be characterized by the abstraction level of the commanded signal, ranging from full-body motion states to task-space targets.
Motion-tracking and teleoperation pipelines often retarget human whole-body motion to humanoid references in joint or keypoint space, enabling high-fidelity imitation as well as large-scale demonstration collection \cite{twist2}.
To broaden task coverage, multi-mode controllers combine locomotion-oriented root commands (e.g., velocity and height) with upper-body tracking signals (joint- or keypoint-based), allowing the interface to switch among multiple command spaces according to task context \cite{hover}. A complementary line of work exposes more task-level, interpretable targets such as end-effector (EE) poses or sparse keypoints, obtained from VR inputs or visual perception, and executes them through a low-level whole-body tracker \cite{sonic,visualmimic,hero}.
Such interfaces naturally support hierarchical system designs in which a planner or mid-level policy generates trajectories or waypoint sequences, while a tracking controller realizes them on the robot.
Recent compliance-oriented interface designs further motivate decoupled and keypoint-based abstractions for interaction scenarios, reducing the burden of specifying full-body motion states \cite{chip}.
Motivated by these trends, our method adopts a compact end-effector--root (EE-root) command space that specifies root motion together with EE pose targets, providing a unified interface for both teleoperation and hierarchical planning in contact-rich humanoid loco-manipulation.

\vspace{-2mm}
\section{Method}

\begin{figure*}[htbp]
\begin{center}
\includegraphics[scale=0.55]
{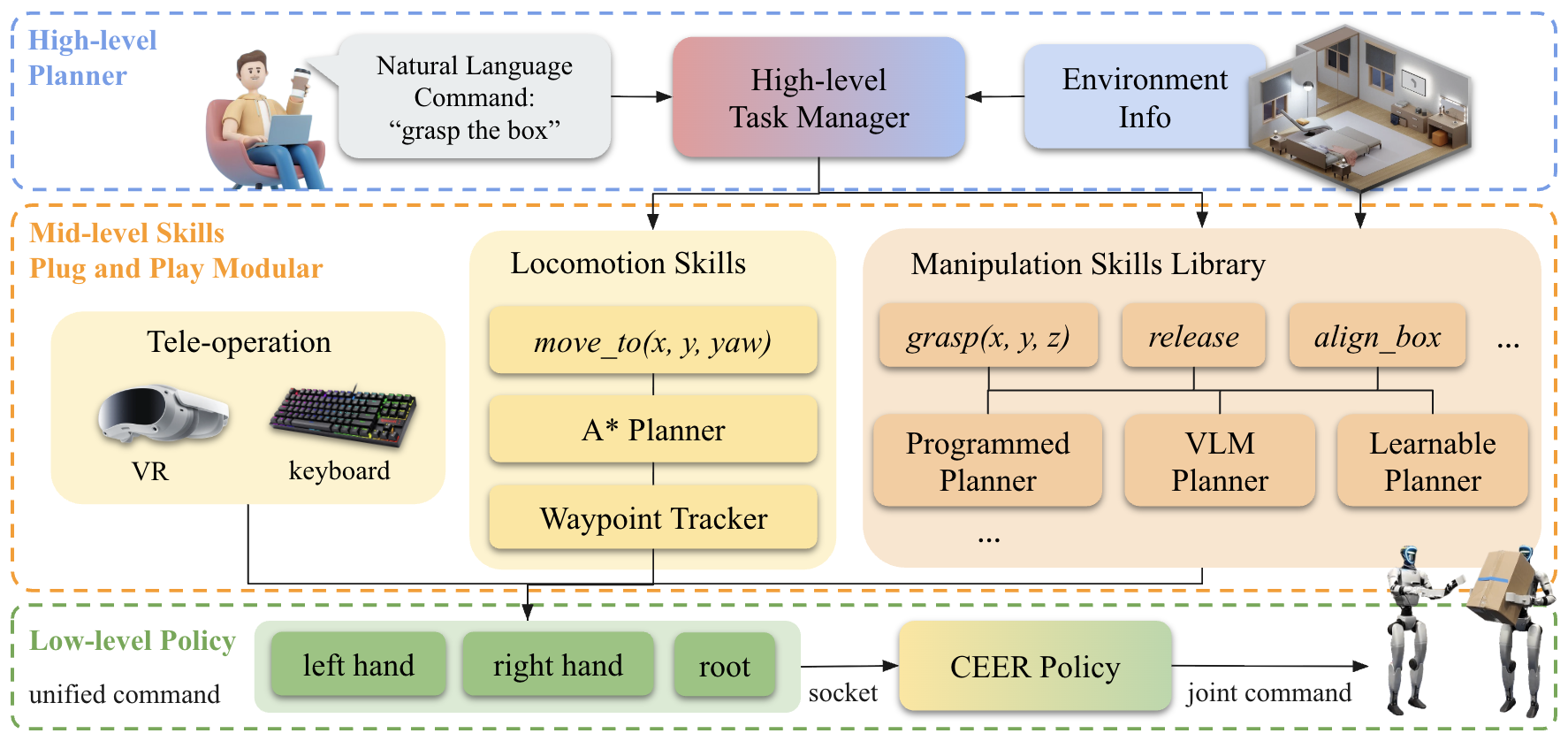}
\end{center}
\vspace{-3mm}
\caption{
Overview of the proposed three-layer hierarchical system. 
At the high level, a language instruction is interpreted by an LLM-based skill manager, which selects and composes mid-level skills based on environmental information. 
The mid-level consists of plug-and-play locomotion and manipulation modules that generate unified end-effector and root commands. 
At the low level, a unified CEER policy converts these commands into joint-space actions. 
This modular design decouples task reasoning, skill execution, and low-level control, enabling scalable and extensible system integration.
}
\label{fig: system framework}
\vspace{-5mm}
\end{figure*}

\subsection{Impedance Control Formulation}

We model robot–environment interaction using a Cartesian impedance formulation:
\begin{equation}
F = K_m(\ddot{x}-\ddot{x}_d) + K_d(\dot{x}-\dot{x}_d) + K_p(x-x_d),
\end{equation}
where $x$ denotes the task-space state, $x_d$ is the desired reference, and $K_m, K_d, K_p$ are inertia, damping, and stiffness gain matrices. 
In practice, we assume zero desired acceleration ($\ddot{x}_d=0$) and constant gains.

Following~\cite{gentlehumanoid}, external interaction is modeled as a spring-like disturbance $f_{ext}$. 
Under quasi-dynamic assumptions (negligible acceleration), the steady-state impedance dynamics reduce to
\begin{equation}
K_p (x_t^{imp} - x_t^{ref}) + K_d (\dot{x}_t^{imp} - \dot{x}_t^{ref}) = f_{ext,t}.
\end{equation}

Assuming velocity tracking is preserved ($\dot{x}_t^{imp}=\dot{x}_t^{ref}$), the compliant target becomes
\begin{equation}
x_t^{imp} = x_t^{ref} + K_p^{-1} f_{ext,t}.
\end{equation}

The impedance-consistent target $(x_t^{imp}, \dot{x}_t^{imp})$ is used as privileged observation and for reward construction during training.

\subsection{Policy Learning}
\label{sec:policy}

\begin{figure}[htbp]
\begin{center}
\includegraphics[scale=0.45]
{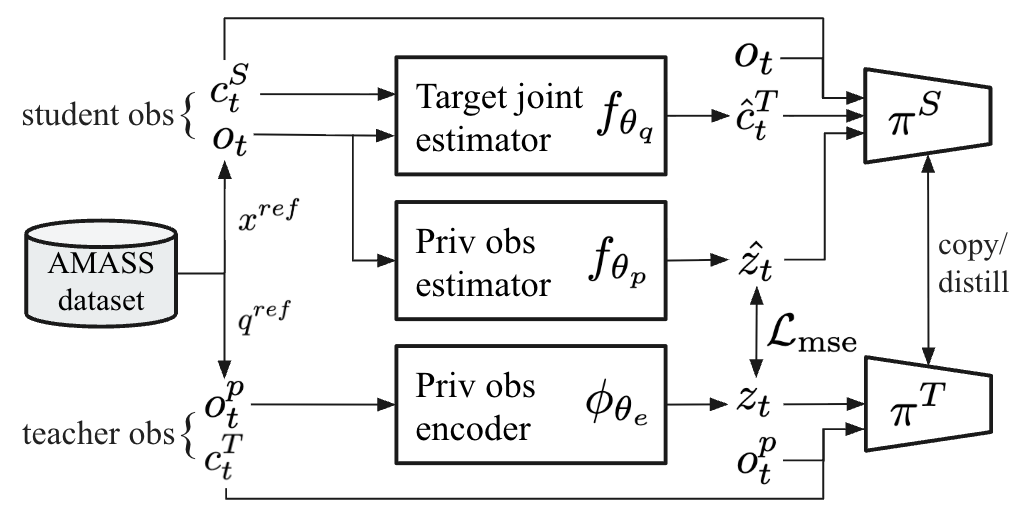}
\end{center}
\vspace{-3mm}
\caption{
Network architecture. We introduce an additional target joint estimator $f_{\theta_q}$ that predicts the full-body joint reference from the student observation $(o_t, c_t^{S})$. 
This design aligns the input structure of the student policy with that of the teacher policy, enabling direct weight initialization and stable fine-tuning.
}
\label{fig: network structure}
\vspace{-5mm}
\end{figure}

We adopt motion imitation as the foundation of policy learning. 
A general motion-tracking teacher policy $\pi_T$ is first trained on human motion datasets, and subsequently distilled into a student policy $\pi_S$ conditioned only on EE-root commands. 
Rather than tracking full joint targets, the student implicitly generates coherent whole-body motions from high-level EE-root inputs while retaining the stability of the teacher.

\subsubsection{Observation}

The student policy receives an observation $o_t^{S} = \big( c_t^{S}, \; o_t \big)$, where the command $c_t^{S} \in \mathbb{R}^{16}$ consists of the task-space reference including the root pose $(x, y, z, \text{yaw})$ and the end-effector position and orientation. 
The shared policy observation $o_t$ contains only proprioceptive signals and their histories, which are available on the real robot. Specifically, $o_t = \big( g_t, \omega_t, q_{t-K:t}, a_{t-L:t-1} \big)$ where $g_t$ denotes the projected gravity vector, $\omega_t$ the root angular velocity, $q_{t-K:t}$ the joint positions over the past $K$ steps, and $a_{t-L:t-1}$ the previous $L$ actions.

The teacher policy observes $o_t^{T} = \big( c_t^{T}, \; o_t, \; o_t^{p} \big)$, where the joint-space reference command $c_t^{T} \in \mathbb{R}^{29}$ corresponds to the target joint configuration. In addition to the policy observation $o_t$, the teacher has access to privileged information $o_t^{p} = \big( f^{ext}_{t}, \; s_t^{ref}, \; s_t^{sim}, \; s_t^{act} \big)$, where $f^{ext}_{t}$ denotes the external force applied to the robot, 
$s_t^{ref}$ represents the privileged tracking state derived from the impedance control formulation, 
$s_t^{sim}$ denotes the simulator ground-truth root and joint states, and $s_t^{act}$ corresponds to the actuator states including position and torque.



\subsubsection{Teacher–Student Framework}

Fig.~\ref{fig: network structure} illustrates the overall training architecture of our policy. Following prior teacher–student training paradigms \cite{gentlehumanoid, facet}, we directly provide the joint-space reference command $c_t^{T}$ to the teacher policy. 
Importantly, this signal bypasses the privileged observation encoder and is fed into the teacher policy without latent compression. 
The motivation behind this design is that the reference joint configuration is the primary supervision signal for motion tracking. 
Passing it through an encoder may attenuate its direct influence on policy learning, which could degrade the performance of the teacher policy.

For the student policy, we introduce two separate estimators: 
a privileged state estimator $f_{\theta_p}$ and a target joint estimator $f_{\theta_q}$. 
The privileged state estimator $f_{\theta_p}$ predicts the remaining privileged information based solely on the shared policy observation $o_t$. 
The target joint estimator $f_{\theta_q}$ predicts the 29-dimensional reference joint configuration from the student observation $(o_t, c_t^{S})$.

This design ensures that the student policy $\pi^{S}$ receives inputs with the same dimensionality and structural form as the teacher policy $\pi^{T}$. 
As a result, the student can directly inherit the weights of the teacher policy, which significantly stabilizes and accelerates the fine-tuning process. We adopt a two-stage teacher–student training procedure based on PPO \cite{facet}. 
Both teacher and student policies output joint commands $a_t \in \mathbb{R}^{29}$.


\subsection{Reward Function}

Our reward design largely follows prior work~\cite{gentlehumanoid}, with an additional end-effector tracking term to explicitly encourage task-space accuracy:
\[
r_t^{ee} = \exp\big( -\| x_t^{ee} - x_t^{ref} \|^2 \big),
\]
where $x_t^{ee}$ denotes the current end-effector pose and $x_t^{ref}$ the reference target. The total reward is defined as:
\[
r_{\text{total}} =
w_{ee} r_{ee}
+ w_c r_{\text{compliance}}
+ w_m r_{\text{motion}}
+ w_l r_{\text{locomotion}},
\]
where $w_{ee}, w_c, w_m, w_l$ are scalar weighting coefficients, $r_{\text{compliance}}$ promotes compliant interaction behavior, 
$r_{\text{motion}}$ encourages faithful motion tracking, 
and $r_{\text{locomotion}}$ stabilizes lower-body locomotion.

\subsection{System Integration}

In this section, we demonstrate how the proposed low-level EE-root policy serves as an interpretable and planner-compatible interface within a hierarchical control architecture. 
We first introduce the overall three-layer system framework shown in Fig.~\ref{fig: system framework}, followed by the design of representative mid-level skills and the high-level task manager.

\subsubsection{Three-Layer Hierarchical System}

As illustrated in Fig.~\ref{fig: system framework}, we construct a three-layer hierarchical architecture consisting of a low-level whole-body controller, a modular mid-level skill layer, and a high-level task manager.

\textbf{Low-Level Control.} 
The compliant EE-root control policy trained in Section~\ref{sec:policy} serves as the low-level whole-body controller in our hierarchical architecture. 
CEER consumes a standardized task-level command representation defined in the EE-root space, which specifies root motion and end-effector pose targets. 
Instead of tracking joint trajectories, CEER directly operates in this interpretable command space and maps EE-root commands to dynamically feasible joint-level PD targets for execution. 

The controller runs at 50\,Hz, ensuring stable and responsive whole-body tracking under contact. 
By defining control in the EE-root space, CEER decouples task specification from morphology-specific joint targets, enabling compatibility with heterogeneous mid- and high-level planners.

\textbf{Mid-Level Skill Layer.} 
The mid-level layer consists of modular, plug-and-play skills that are invoked by the high-level task manager. 
Each mid-level module outputs commands formatted in the unified EE-root space, making them directly compatible with the low-level controller. 
Typical examples include teleoperation interfaces (VR and keyboard), programmed locomotion primitives (e.g., move-to-position with yaw control), and manipulation primitives such as grasp and place. 
Learning-based skills (e.g., reinforcement learning, imitation learning, or diffusion-based policies) can also be integrated at this layer without modifying the low-level controller. 
These modules typically operate at 10--20\,Hz.
Importantly, if the low-level controller were defined in joint space, many mid-level policies would either become incompatible or require retraining, since most planners naturally generate waypoints or end-effector targets rather than full joint trajectories. 
The EE-root abstraction therefore enables flexible integration of heterogeneous skill modules.

\textbf{High-Level Task Manager.} 
At the top layer, an LLM functions as a task manager, decomposing instructions into mid-level skill calls on a second-level time scale. This architecture ensures high-level planning remains decoupled from low-level dynamics via the unified EE-root interface.

\subsubsection{Middle-level Skills}
\label{subsubsec:Middle-level Skills}
We categorized mid-level skills as teleoperation, locomotion and manipulation. Together, they demonstrate that heterogeneous planners can be seamlessly connected through the unified EE-root interface.


\textbf{Teleoperation Skills.} We implement two teleoperation interfaces: keyboard and VR. 
All teleoperation modules generate commands directly in the EE-root space, demonstrating the interpretability of the proposed interface. For VR-based control, the 6-DoF pose of each hand is obtained from the VR controllers in the world frame. 
These poses are transformed into the robot coordinate frame and directly used as end-effector targets. In addition, we provide a lightweight keyboard teleoperation script for rapid testing. For symmetric grasping tasks, we define a simple rule-based mapping: the two end-effectors share identical targets in the forward ($x$) and vertical ($z$) directions, while the lateral ($y$) component is mirrored to maintain symmetric arm motion. 



\textbf{Locomotion Skill.} Our locomotion skill is based on a path planning method that converts high-level navigation goals into EE-root commands. 
Given a grid-based map and a target $(x, y, \theta)$ pose, a hybrid A* planner generates a sequence of waypoints. 
The planner is adapted to humanoid constraints, allowing in-place rotation and backward motion when necessary. A waypoint tracker converts the planned path into root velocity commands, while the end-effector targets remain unchanged. 

\textbf{Manipulation Skills.} We implement three types of manipulation skills that are planned, vision language model(VLM)-based, or learned. The variety of such skills demonstrate the flexibility of the EE-root abstraction. 
The planned skills are usually derived from parameterized waypoints, rule-based trajectory generation, or classical motion planners. 
These waypoints are directly mapped to EE-root commands.
The VLM skills take egocentric visual observations and natural language instructions to predict an end-effector trajectory for task completion. 
The predicted waypoints are transformed into the robot frame via hand–eye calibration before execution.
For learned skills, we deploy a diffusion-based imitation learning pipeline. 
By aligning all planner outputs to the EE-root command format, new manipulation skills can be integrated without modifying or retraining the low-level controller.


\subsubsection{High-level task manager}

\begin{figure*}[htbp]
\begin{center}
\includegraphics[scale=0.28]
{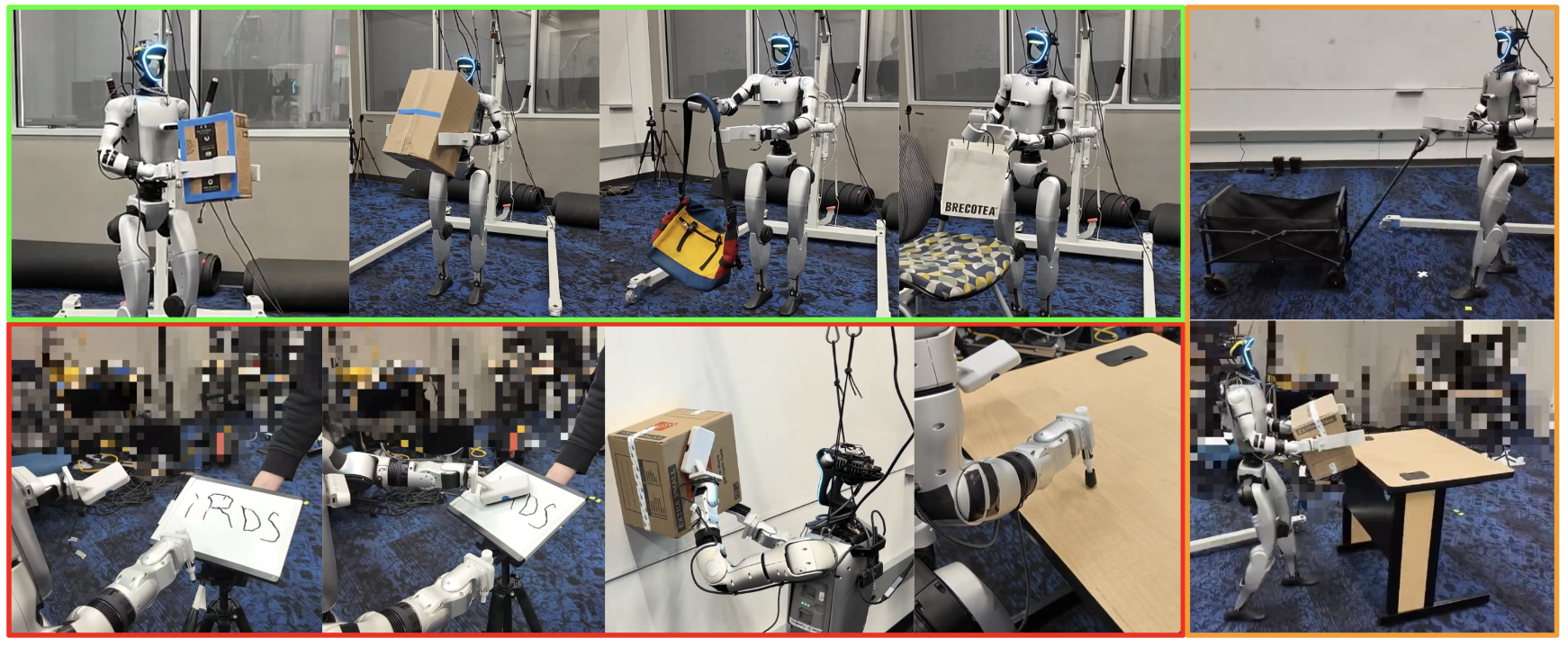}
\end{center}
\vspace{-3mm}
\caption{
Real-world tele-operation evaluation across task categories. 
\textcolor{green}{\rule{0.8em}{0.8em}}~Simple manipulation: small and large box grasping, single-shoulder bag handover, and paper bag handover. 
\textcolor{red}{\rule{0.8em}{0.8em}}~Forceful manipulation: writing, wiping, box-on-wall rotation, and pen cap insertion. 
\textcolor{orange}{\rule{0.8em}{0.8em}}~Loco-manipulation: cart pulling and box transportation.
}
\label{fig: exp2a}
\vspace{-5mm}
\end{figure*}

We build an LLM-based task manager that decomposes natural language instructions into sequences of mid-level skill invocations. Given a command such as ``pick up the box from the table and place it on the bed,'' the agent first combines the instruction with a system prompt encoding domain knowledge and sends it to the LLM as the initial conversation context. The LLM then selects a skill to execute through a function-calling interface, where each skill is registered as a typed function with spatial parameters: locomotion skills accept a target pose $(x, y, \theta)$; end-effector skills accept hand positions in the robot body frame; and compound skills such as grasp and release encapsulate multi-step manipulation sequences. The selected skill is dispatched to the corresponding mid-level module for execution. After each execution, the agent queries the current robot state, including root pose, end-effector positions, end-effector safety status, and detected object locations, and appends the observation to its conversation history. Based on the updated context, the LLM decides whether to invoke the next skill or signal task completion. This closed-loop cycle of reason--act--observe~\cite{react} repeats until the task is finished, allowing the agent to verify action outcomes, recover from execution errors, and adapt to environmental changes without generating a complete plan upfront.

Rather than fine-tuning the LLM for robot control, we encode manipulation expertise directly in the system prompt. The prompt specifies the workspace coordinate system and spatial layout, canonical hand position states (\textsc{rest}, \textsc{hold}, \textsc{ready}, \textsc{grasp}) with safety classifications indicating whether each configuration permits locomotion, and recommended action sequences for common loco-manipulation tasks such as pick-and-place. This contextual knowledge enables the LLM to reason about spatial goals and safety constraints without task-specific training.

The feasibility of this LLM-agent integration rests on the interpretability of the EE-root interface. Because all mid-level skills, whether planned, learned, or LLM-scheduled, produce commands in the same end-effector and root-velocity space, the LLM reasons purely about spatial goals and action sequencing without exposure to joint-level dynamics. The domain knowledge in the system prompt can be expressed entirely in terms of end-effector positions and root poses, both of which are physically intuitive quantities. This shared abstraction is what allows a general-purpose language model to serve as a task-level controller for a full humanoid system.

\section{Experiment}

We conduct comprehensive experiments to evaluate both the control performance of CEER and the scalability of the hierarchical system. 
Specifically, we investigate the following research questions:

\begin{itemize}
    \item How accurately does CEER track end-effector position and orientation?
    \item Does compliant control enable stable contact-rich dual-arm manipulation under teleoperation?
    \item To what extent can heterogeneous mid-level planners be integrated through the EE-root interface without retraining the low-level controller?
    \item Can the hierarchical system execute language-specified long-horizon tasks in a room-scale environment, and how does it compare to human teleoperation?
\end{itemize}

\subsection{Experimental Setup}

All real-world experiments are conducted on a Unitree G1 humanoid robot equipped with a RealSense D435 egocentric camera. To support different manipulation tasks, we design three lightweight and easily attachable end-effector tools, including a hook manipulator, a pen holder, and a whiteboard eraser holder. The CEER policy is trained in simulation using Isaac Lab with 16,384 parallel environments on four NVIDIA L40s GPUs for 8 hours. For deployment, the policy runs on a desktop workstation equipped with an Intel Core Ultra 9 CPU and an NVIDIA RTX 5090 GPU, and is executed on the robot through the RoboJudo pipeline~\cite{robojudo}.

For real world teleoperation experiments, we use a Pico 4 VR headset. ALVR and SteamVR are employed to establish communication between the Ubuntu 24.04 workstation and the VR device via a wired connection to ensure low-latency control.

\subsection{Tracking performance of low-level policy}

\begin{table}[t]
\centering
\caption{Tracking performance evaluation.}
\label{tab:tracking_results}
\resizebox{\columnwidth}{!}{
\begin{tabular}{lccc}
\toprule
\textbf{Method} 
& $\epsilon_p$ (m) 
& $\epsilon_r$ (rad) 
& RMS($j$) (m/s$^3$) \\
\midrule
Ours & 0.033 (0.011) & 0.320 (0.093) & 4.5e3 \\
Ours w/o $f_{\theta_q}$ & 0.124 (0.521) & 0.9 (0.494) & 3.3e4 \\
No-force RL & 0.028 (0.010) & 0.317 (0.110) & 9.9e3 \\
End-to-end RL & 0.053 (0.184) & 0.349 (0.081) & 6.9e3 \\
\bottomrule
\end{tabular}
}
\vspace{-5mm}
\end{table}


We evaluate the end-effector tracking performance of the proposed CEER policy in simulation. 
We compare against three baselines: (1) \textit{Ours w/o $f_{\theta_q}$}, which adopts a conventional teacher–student framework without the joint target estimator. (2) \textit{No-force RL}, which follows the same training pipeline as CEER but removes external force disturbances during training. This baseline represents obtaining an EE-root controller through motion-tracking-based learning without compliance modeling. (3) \textit{End-to-end RL}, modified from SkillBlender \cite{skillblender}, which learns the EE-root controller directly from task rewards without motion imitation. Together, these baselines represent two common approaches for learning task-space humanoid controllers: motion imitation and end-to-end reinforcement learning.
Results are summarized in Table~\ref{tab:tracking_results}. 
We report the root mean squared error (RMSE) and standard deviation of end-effector position and orientation errors ($\epsilon_p$ and $\epsilon_r$), as well as the root mean square of jerk (RMS($j$)) to quantify motion smoothness. 

In absolute terms, our method achieves a position RMSE of 3.3\,cm and orientation error of 0.32\,rad, while significantly reducing RMS jerk to $4.5\times10^3$, indicating accurate and smooth task-space control.
In relative terms, our method achieves higher tracking accuracy than both \textit{End-to-end RL} and \textit{Ours w/o $f_{\theta_q}$}, while attaining comparable accuracy to \textit{No-force RL}. 
This is expected, as removing disturbance during training simplifies the learning objective and can lead to slightly lower tracking error. 
Importantly, the performance gap between CEER and \textit{Ours w/o $f_{\theta_q}$} demonstrates that the joint target estimator is critical for learning a high-quality tracking policy. Notably, CEER achieves the lowest jerk among all methods, indicating substantially reduced high-frequency oscillations. 
This suggests that the compliant design produces smoother and more stable end-effector motions, as the impedance-based formulation inherently dampens high-frequency error corrections and mitigates abrupt force responses.
Consistent with this quantitative result, we observe visibly reduced shaking in simulation compared to the other policies.



\subsection{Teleoperation for Real-world Tasks}
We evaluate the real-world contact-rich manipulation capabilities enabled by CEER’s compliant control through teleoperation experiments across three categories of tasks: (1) simple manipulation, (2) forceful manipulation, and (3) loco-manipulation. Fig. \ref{fig: exp2a} illustrates the three task categories, distinguished by colored bounding boxes; corresponding live experiments are provided in the supplementary video.

\textbf{\textcolor{green}{\rule{0.8em}{0.8em}} Simple Manipulation.} 
These tasks involve basic object contact and grasping. 
We evaluate grasping boxes of different sizes and perform object handover using a single-shoulder bag and a paper bag. 
The robot is able to transfer objects between hands and open the paper bag using dual-handle grasping without damaging it. 
These results demonstrate stable dual-arm coordination under compliant control.

\textcolor{red}{\rule{0.8em}{0.8em}} \textbf{Forceful Manipulation.} 
These tasks require more accurate contact regulation and force modulation. 
In the whiteboard writing and wiping experiment, the robot is equipped with a pen holder on the right hand and an eraser holder on the left hand. 
The robot is able to write human-recognizable letters (“IROS”) in two out of three attempts, and wiping can be performed consistently. For the box-on-wall rotation task, the left end-effector presses the box against the wall to generate sufficient friction while the right end-effector applies torque using a hook attachment. The robot successfully completes six consecutive rotations without dropping the object, demonstrating stable multi-contact manipulation. In the pen-cap insertion task, the robot inserts the pen tip into a tight pen cap, which requires precise alignment and force application. Although the cap mechanism is mechanically stiff, the robot is able to achieve consistent insertion under teleoperation.

\textbf{\textcolor{orange}{\rule{0.8em}{0.8em}} Loco-Manipulation.} 
For loco-manipulation tasks, the robot is controlled via keyboard input. 
In the cart-pulling task, the robot grasps and pulls an empty cart using a hook attachment. 
In the box transportation task, the robot grasps a box from a table, moves backward and forward, and restores the object to its original position.

Overall, the compliant nature of CEER enables safe and stable contact during teleoperation, while maintaining sufficient tracking accuracy for complex manipulation. 
Locomotion behaviors are functional but less natural and smooth compared to manipulation, suggesting that additional locomotion-specific training data may further improve performance.

\subsection{Planner Compatibility Study}

\begin{figure}[htbp]
\vspace{-3mm}
\begin{center}
\includegraphics[scale=0.34]
{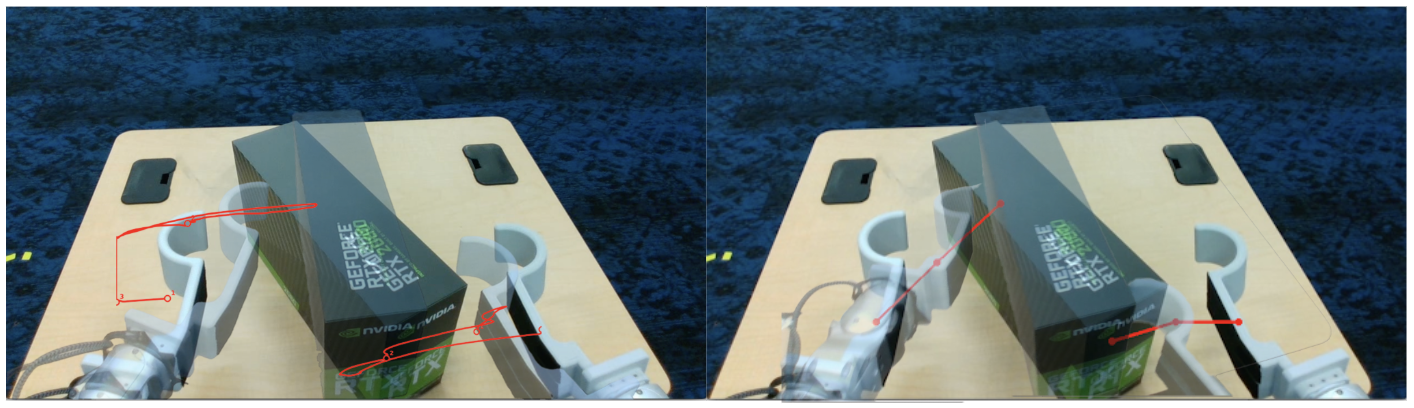}
\end{center}
\vspace{-2mm}
\caption{End-effector trjectory generated by diffusion policy (left) and VLM (right).}
\label{fig: exp2b}
\vspace{-3mm}
\end{figure}

This section demonstrates the planner-agnostic property of the proposed EE-root abstraction. We focus on verifying the interface compatibility as qualitative validation. 
We integrate the three proposed heterogeneous mid-level planners (detailed in Sec. \ref{subsubsec:Middle-level Skills}) into a unified low-level CEER controller:
a rule-based programmed planner, a diffusion-based planner, and a VLM planner. Each planner produces task-level trajectories in its native representation, which are aligned to the unified EE-root command space and executed by the same low-level policy without retraining or architectural modification. Fig.~\ref{fig: exp2b} visualizes representative end-effector trajectories generated by the diffusion-based and VLM planners. Despite their heterogeneous internal structures, both planners produce compatible EE-root commands that can be directly consumed by CEER.





\subsection{System evaluation}

\begin{figure}[htbp]
\vspace{-3mm}
\begin{center}
\includegraphics[scale=0.13]
{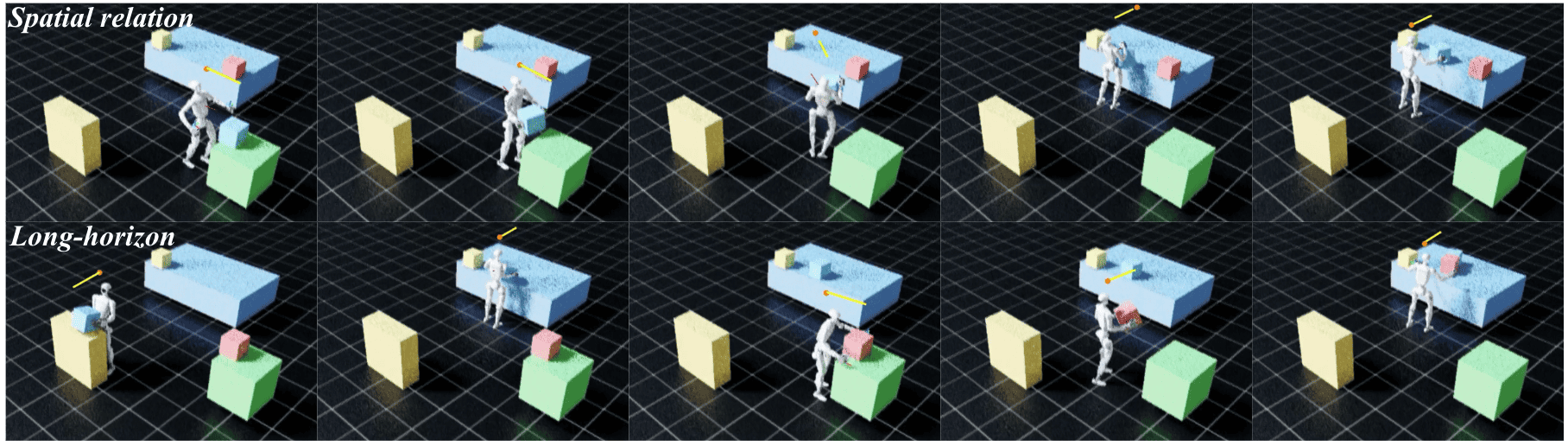}
\end{center}
\vspace{-3mm}
\caption{Screenshots of spacial relation and long-horizon tasks in simulation. }
\label{fig:exp3}
\vspace{-3mm}
\end{figure}

We evaluate the capability and connectivity of the proposed three-layer hierarchical system in simulation. 
Our hypothesis is that, under a unified EE-root interface, household tasks can be decomposed into locomotion, manipulation invocation, and task completion. 
To isolate system-level behavior rather than skill diversity, we intentionally adopt a minimal skill set and use a single grasp primitive for all tasks.

\paragraph{Environment Setup}
The simulated room contains a bed, sofa, and table with randomized planar positions ($\pm 0.5$\,m) and heights ($0.5–0.7$\,m). 
Three boxes with randomized initial positions ($\pm 0.25$\,m) are included. 
The robot starts at the room center.

\paragraph{Task Suite}
We evaluate three task categories of increasing complexity:
(1) \textit{Simple tasks}: arm motion and navigation without object interaction, evaluated under two instruction modes: explicit commands and paraphrased natural language variations. (20 trials)
(2) \textit{Single-object loco-manipulation}: 
    We further divide this category into three subtypes:
    (a) Explicit placement tasks, 
    (b) Linguistic variation tasks, and 
    (c) Spatial-relation tasks (30 trials).
(3) \textit{Long-horizon tasks}: multi-object transport requiring sequential skill invocation (10 trials).

\paragraph{Metrics}
We report success rate and average completion steps and time.

\paragraph{Human Baseline}
Five participants perform the same tasks via keyboard teleoperation with the same control degrees of freedom as the robot. After practice, each participant is given two attempts per task type (40 trials total).
We record success rate and completion time.

\begin{figure}[t]
\vspace{1mm}
\centering

\begin{minipage}{0.48\columnwidth}
\centering
\captionof{table}{Performance on simple tasks.}
\label{tab:simple}
\resizebox{\linewidth}{!}{
\begin{tabular}{lcc}
\toprule
Task & Explicit & Paraphrased \\
\midrule
Arm Motion  & 5/5 & 4/5 \\
Navigation  & 5/5 & 5/5 \\
\bottomrule
\end{tabular}
}
\end{minipage}
\hfill
\begin{minipage}{0.46\columnwidth}
\centering
\includegraphics[width=\linewidth]{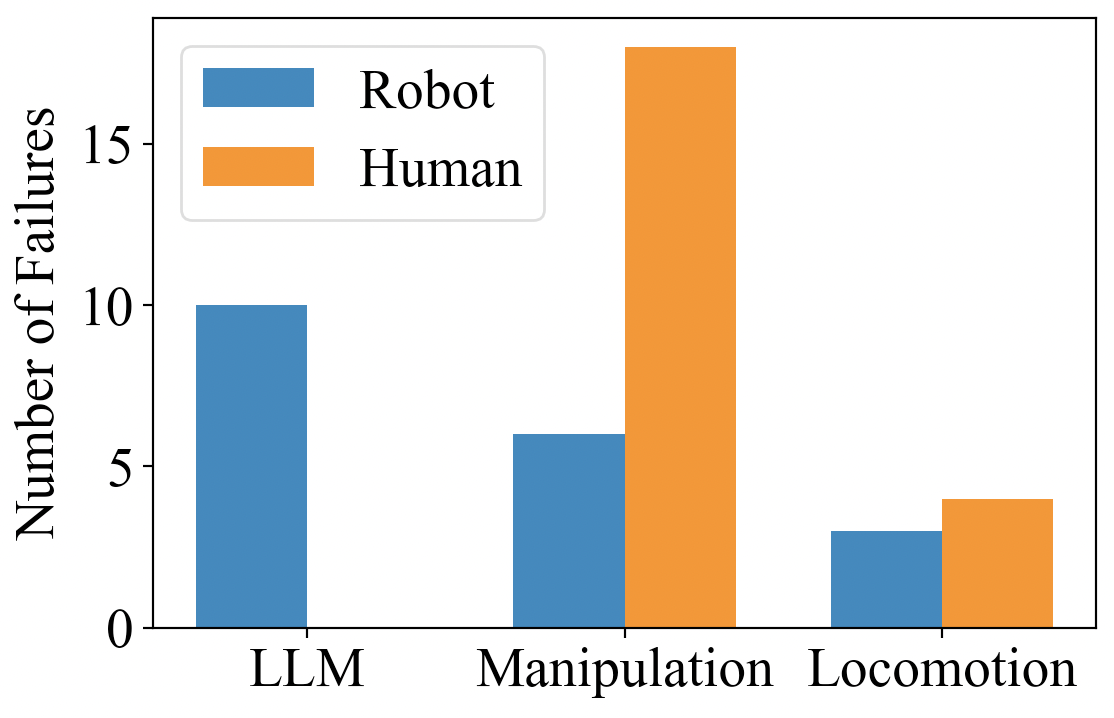}
\captionof{figure}{Failure analysis of robot and human.}
\label{fig:failure}
\end{minipage}
\vspace{-7mm}
\end{figure}

\paragraph{Results}
Results are summarized in Table  \ref{tab:system_eval}, \ref{tab:simple}, with representative rollouts shown in Fig.~\ref{fig:exp3}. 


\begin{table}[t]
\vspace{2mm}
\centering
\caption{System-level evaluation and comparison with human tele-operation.}
\label{tab:system_eval}
\resizebox{\columnwidth}{!}{
\begin{tabular}{lcccccc}
\toprule
\textbf{Task Category} 
& \multicolumn{3}{c}{\textbf{Ours}} 
& \multicolumn{2}{c}{\textbf{Human}} \\
\cmidrule(lr){2-4} \cmidrule(lr){5-6}
& Success & Steps & Time (s) 
& Success & Time (s) \\
\midrule
Explicit Placement & $6/10$ & $15.2$ & $72.53$ & $3/10$ & $118.67$ \\
Linguistic Variation  & $7/10$ & $14.7$ & $72.16$ & $4/10$ & $89.75$ \\
Spatial Relation & $7/10$ & $12.8$ & $80.72$ & $7/10$ & $240.43$ \\
\midrule
Long-Horizon (2-Obj) & $2/10$ & $35.5$ & $166.49$ & $4/10$ & $232.25$ \\
\bottomrule
\end{tabular}
}
\vspace{-5mm}
\end{table}

The system achieves near-perfect performance on simple tasks, with the only failure caused by incorrect LLM skill invocation. 
For single-object tasks, performance is comparable to human teleoperation while requiring shorter completion time, indicating reliable coordination between navigation and manipulation. For long-horizon multi-object tasks, human operators achieve 4 successful trials out of 10, compared to 2 out of 10 for the autonomous system.

Fig \ref{fig:failure} shows the failure cases of both human and robot. All failure cases are categorized into LLM errors (misunderstanding or incorrect parameters), manipulation failures (grasp/release errors), and locomotion failures (object drop during transport). 
$52.6\%$ of autonomous failures stem from LLM-level planning inaccuracies in estimating orientation angles and distances, resulting in unstable grasps that underscore the necessity for robust spatial reasoning.

Interestingly, humans exhibit more manipulation failures under the same control interface, likely due to limited force feedback. In contrast, the robot leverages a grasping primitive that ensures higher consistency, potentially explaining its superior grasping performance relative to human operators.
However, in long-horizon tasks, humans demonstrate stronger recovery ability, compensating through more timely and reactive adjustment despite longer execution time.

Overall, CEER enables reliable execution for moderate-complexity tasks, while long-horizon performance remains bounded by planning accuracy and recovery mechanisms.

\section{Conclusions}

We presented CEER, a compliant EE-root control abstraction for humanoid loco-manipulation. By defining task specifications through root motion and EE-pose targets, CEER decouples task-level planning from joint-level control while maintaining compliant whole-body behavior. This unified interface enables a hierarchical system for plug-and-play integration of heterogeneous skills without retraining the low-level policy. Extensive simulation and hardware experiments demonstrate CEER’s efficacy in contact-rich manipulation and long-horizon tasks, highlighting its potential as a modular interface for scalable humanoid intelligence.


The current implementation has several limitations. First, due to the absence of an adaptive stiffness modulation mechanism, the impedance parameters are fixed during execution. Second, while the policy supports basic locomotion, end-effector targets must be explicitly provided. The arms remain static unless swing trajectories are manually specified. 



Future work will incorporate adaptive impedance control for task-dependent stiffness modulation and extend the policy input space to support flexible contact specifications, such as elbow or foot placement. Furthermore, we aim to enable partial target specification—e.g., providing only root commands during walking—to facilitate more autonomous and natural locomotion behaviors.

\addtolength{\textheight}{-12cm}   






\bibliographystyle{IEEEtran}
\bibliography{references}

\end{document}